\documentclass[letterpaper, 10 pt, conference]{ieeeconf}  

\IEEEoverridecommandlockouts                              

\usepackage{graphics} 
\usepackage{epsfig} 
\usepackage{times} 
\usepackage{amsmath} 
\usepackage{amssymb}  

\usepackage{hyperref}
\hypersetup{
    colorlinks=true, 
    linkbordercolor={1 0 0}, 
    pdfborderstyle={/S/U/W 1} 
}
\usepackage{multirow}  
\usepackage{caption}   
\usepackage{subcaption}
\usepackage{booktabs}
\usepackage{cite} 
\usepackage{wrapfig} 

\title{\LARGE \bf
StreamMOTP: Streaming and Unified Framework for Joint 3D Multi-Object Tracking and Trajectory Prediction
}

\author{Jiaheng Zhuang$^{1}$, Guoan Wang$^{2}$, Siyu Zhang$^{2}$, Xiyang Wang$^{2}$, \\ Hangning Zhou$^{2}$, Ziyao Xu$^{2}$, Chi Zhang$^{2}$, Zhiheng Li$^{1}$
\thanks{$^{1}$ are with Tsinghua University, China.}%
\thanks{$^{2}$ are with Mach Drive, China.}
\thanks{Email:\tt\small zhuangjh22@mails.tsinghua.edu.cn }%
}

\begin{document}

\maketitle
\thispagestyle{empty}
\pagestyle{empty}

\begin{abstract}

3D multi-object tracking and trajectory prediction are two crucial modules in autonomous driving systems. Generally, the two tasks are handled separately in traditional paradigms and a few methods have started to explore modeling these two tasks in a joint manner recently. However, these approaches suffer from the limitations of single-frame training and inconsistent coordinate representations between tracking and prediction tasks. In this paper, we propose a streaming and unified framework for joint 3D Multi-Object Tracking and trajectory Prediction (StreamMOTP) to address the above challenges. Firstly, we construct the model in a streaming manner and exploit a memory bank to preserve and leverage the long-term latent features for tracked objects more effectively. Secondly, a relative spatio-temporal positional encoding strategy is introduced to bridge the gap of coordinate representations between the two tasks and maintain the pose-invariance for trajectory prediction. Thirdly, we further improve the quality and consistency of predicted trajectories with a dual-stream predictor. We conduct extensive experiments on popular nuSences dataset and the experimental results demonstrate the effectiveness and superiority of StreamMOTP, which outperforms previous methods significantly on both tasks. Furthermore, we also prove that the proposed framework has great potential and advantages in actual applications of autonomous driving.

\end{abstract}

\begin{figure}[!htp]
\centering
\includegraphics[scale=0.48]{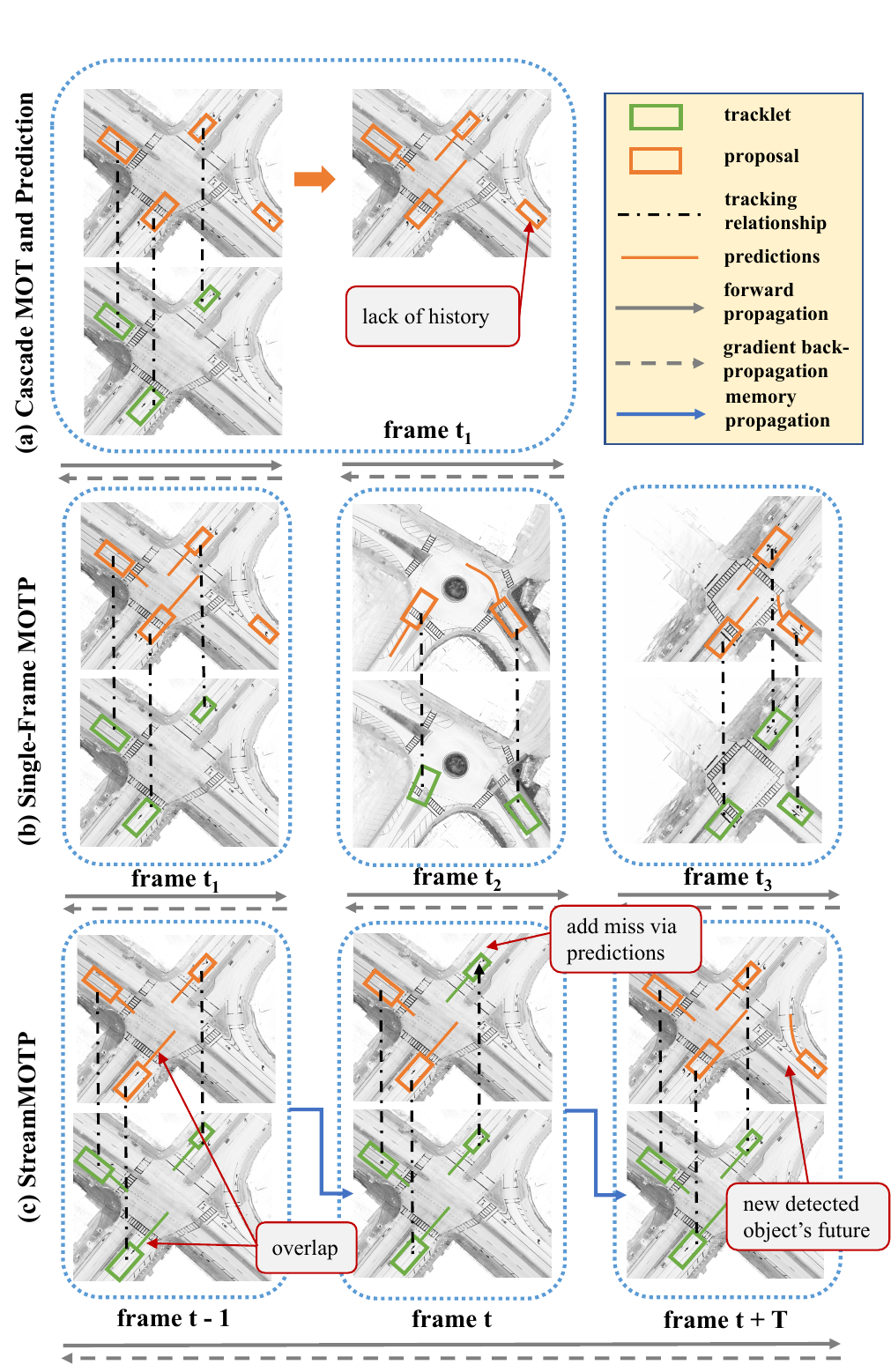}
\caption{Different pipelines for the tasks of multi-object tracking and trajectory prediction in autonomous driving. \textbf{(a)} Cascade paradigm, where the two tasks are performed separately with non-differentiable transitions. \textbf{(b)} Joint single-frame paradigm, where the two tasks are performed jointly in a parallelized framework per frame. \textbf{(c)} The proposed StreamMOTP, where the memory, feature, and gradient are propagated across consecutive frames to enhance the long-term modeling ability and temporal consistency.}
\vspace{-10pt}
\label{fig:stream}
\end{figure}
\section{INTRODUCTION}
In autonomous driving systems, 3D Multi-Object Tracking (MOT) \cite{zhang2019robust, weng2020gnn3dmot, weng2020ab3dmot, bai2022transfusion, wang2023camo, li2023poly} and trajectory prediction \cite{gupta2018social, chandra2019traphic, liang2016semantic, gao2020vectornet, gu2021densetnt, varadarajan2022multipath++, shi2022motion, zhou2023query} are two crucial tasks which play a vital role in ensuring the driving performance of ego-vehicle. Obviously, high-precision tracking can provide a more solid foundation for prediction, and in turn, accurate predictions can enhance the effectiveness of tracking. As depicted in Fig.\ref{fig:stream} (a), the two tasks are executed one after another in current mainstream pipelines of autonomous driving. Although this paradigm has achieved some success, the separated processing flow can not fully exploit the potential complementarity between the tasks of tracking and prediction, since it suffers from information loss, feature misalignment, and error accumulation across modules \cite{weng2022mtp}. Despite some methods \cite{weng2021ptp, weng2022whose, zhang2023towards} attempt to integrate the two tasks as shown in Fig.\ref{fig:stream} (b), some limitations and problems have still not been well explored: (1) the tasks of multi-object tracking and trajectory prediction are both executed in a streaming manner in actual deployments, while the training procedure of most previous methods is conducted in a snap-shot pattern, where the length of historical window is fixed and the long-term information can not be fully exploited efficiently. (2) In general, the coordinates representation of objects for tracking and prediction are different, where a unified coordinate system is needed in MOT for optimal association while most prediction methods adopt the agent-centric coordinate representation for each object to ensure pose-invariance. (3) Most methods focus on predicting the future trajectories of objects visible in current frame, inadvertently overlooking those lost because of either occlusions or miss from upstream perception, which may result in adversely affecting downstream tasks.

In this paper, we introduce \textbf{StreamMOTP}, a streaming framework for joint multi-object tracking and trajectory prediction as depicted in Fig.\ref{fig:stream} (c), where the tasks of MOT and trajectory prediction are jointly performed on successive frames. Specifically, we associate the newly perceived objects with historical tracklets and predict their future trajectories simultaneously. Different from previous works, the extracted latent features of objects are sequentially utilized in StreamMOTP as part of the representation for the subsequent tracked objects during the forward propagation phase. As for the back-propagation, the gradients are not confined to a single frame but are propagated through multiple frames, which greatly narrows the gap between training and online inference, allowing for a more comprehensive learning process by accounting for temporal dependencies across the entire sequence. 

Concretely, we extend the pattern of training from single-frame to multi-frame and introduce a memory bank to maintain and update long-term latent features for tracked objects, thereby improving the model's capability for long-term sequence modeling. Aiming to address the coordinate system discrepancy between the tasks of tracking and prediction, we propose a relative Spatio-Temporal Positional Encoding (STPE) strategy, which is applied to realize the compromise and unification of the different agent- and ego-centric representation in the two tasks. At the same time, based on the observation that there is an obvious overlap between the predicted trajectories of objects in consecutive adjacent frames as depicted in Fig.\ref{fig:stream} (c-left), we apply dual-stream predictor to effortlessly and elegantly generate future trajectories for both tracked and new-come objects simultaneously, which benefits to both tasks of MOT and trajectory prediction. 

It should be pointed out that, with the design of the streaming and unified framework, StreamMOTP obtains the potential and advantages to handle more complex driving scenarios in actual applications. On the one hand, the predicted trajectories for tracked objects could help deal with the problem of occlusions at the current moment by marking the possible positions of obscured targets in the current frame, as shown in Fig.\ref{fig:stream} (c-middle). On the other hand, for the objects newly perceived in the current frame, StreamMOTP maintains the capability to predict their future trajectories by leveraging social interactions and contextual features stored in the memory bank while traditional prediction methods may fail due to the lack of historical information about them, as shown in Fig.\ref{fig:stream} (c-right).

The core contributions are summarized as follows:

\begin{itemize}
    \item We propose StreamMOTP, a joint MOT and trajectory Prediction model based on a streaming framework to bridge the gap between training and actual deployment. A memory bank for tracked objects is introduced in this framework for utilizing long-term features more effectively.
    \item We introduce a spatio-temporal positional encoding strategy to construct the relative relationship between objects in different frames, which reaches the compromise and unification of inconsistent coordinate representation in tracking and prediction.
    \item We design a dual-stream predictor to simultaneously predict the trajectories of objects in both the current and previous frames. The predicted trajectory from the previous frame can further assist in predicting newly perceived objects' trajectories, which achieves better temporal consistency in trajectory prediction.
    \item We get better performance for MOT and trajectory prediction on nuScenes, improving AMOTA by 3.84\% and reducing minADE / minFDE by 0.220 / 0.141.
\end{itemize}



\section{RELATED WORK}

\subsection{3D Multi-Object Tracking}

Existing multi-object tracking paradigms, such as tracking-by-detection (DeepSORT\cite{wojke2017simple}, AB3DMOT\cite{weng2020ab3dmot}), Joint Detection and Embedding learning (FairMOT\cite{zhang2021fairmot}, JDE\cite{wang2020towards}), and joint detection and tracking (Tracktor++\cite{bergmann2019tracking}, YONDTMOT\cite{wang2023you}), typically rely on Kalman filters(KFs) to predict the positions of tracked objects for better-association. Yet, KFs require fine-tuning of parameters and struggle with occlusions (PC3TMOT\cite{wu20213d}, DeepFusionMOT\cite{wang2022deepfusionmot}). In contrast, dedicated prediction tasks can provide superior short-term prediction results for tracking, especially in handling complex scenarios such as occlusions. Therefore, combining the two tasks of multi-object-tracking and trajectory prediction can effectively improve the overall performance of multi-object tracking. This combination not only reduces the dependence on traditional methods like KFs but also enhances the robustness and adaptability of the tracking methods.

\subsection{Trajectory Prediction}

There has been significant progress in trajectory prediction recently. With the use of pooling \cite{gupta2018social}, graph convolution \cite{liang2016semantic}, attention mechanism \cite{shi2022motion} \cite{zhou2023query}, vector-based methods \cite{gao2020vectornet} can efficiently aggregate sparse information in traffic scenes. As the future is uncertain, some works (Multipath++ \cite{varadarajan2022multipath++}, HiVT \cite{zhou2022hivt}) predict multimodal future distribution by decoding a set of trajectories from scene context while others (DenseTNT \cite{gu2021densetnt}) generate multimodal prediction by leveraging anchors. Though these methods greatly improve trajectory prediction, most of them use GT past trajectories as input for training and testing, neglecting tracking error accumulation with imperfect inputs. Therefore, we handle the tasks of tracking and prediction jointly with no need for GT trajectories as predictor's inputs to provide more robust predictions based on practical detectors in the real world.

\begin{figure*}[!htbp]
\centering
\includegraphics[width=\textwidth]{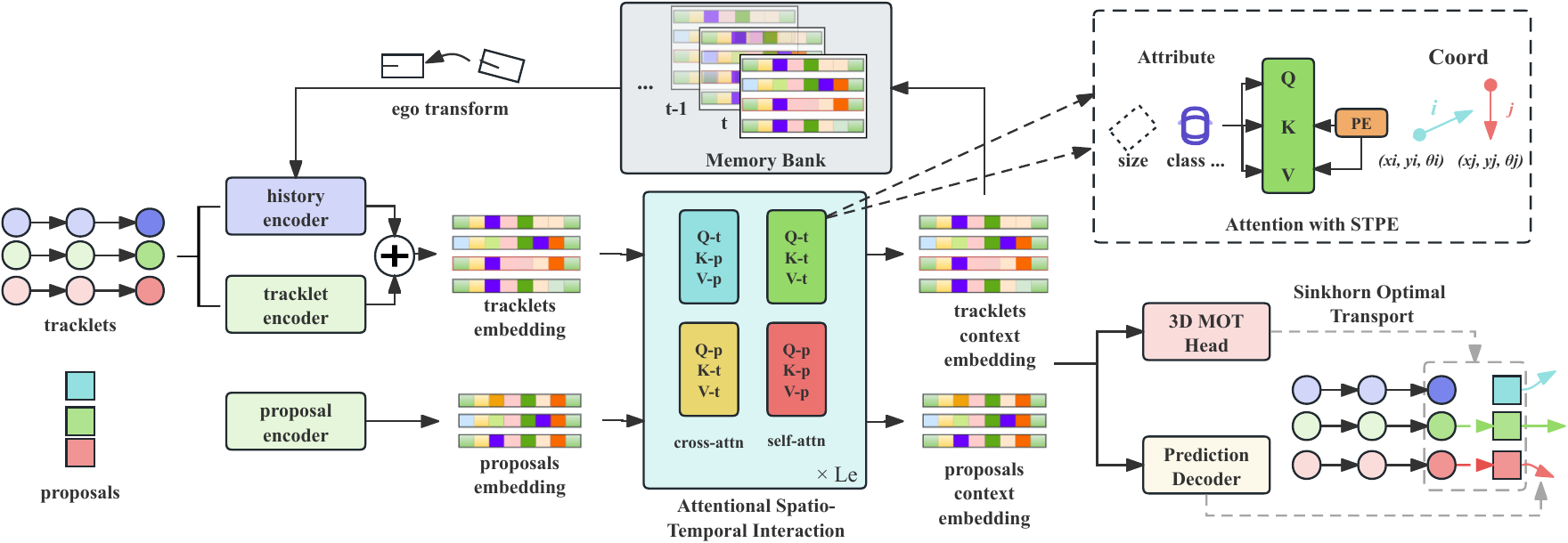}
\caption{Overview of StreamMOTP. Tracklets and proposals denote the previous frame trajectories and the current frame detections respectively. The model first performs Attentional Spatio-Temporal Interaction, which is based on attention with STPE, to get context features. The tasks of tracking and prediction are then performed based on those context features. Memories with up-to-date context features and tracking results are updated at each time step.}
\vspace{-10pt}
\label{fig:models}
\end{figure*}

\subsection{Joint Tracking and Prediction}

In the last couple of years, there has been growing interest in joint tracking and prediction. For example, \cite{yu2021towards} refine the inputs for the predictor through a re-tracking module, MTP \cite{weng2022mtp} propose multi-hypothesis data association to generate multiple sets of tracks for predictor simultaneously. Besides polishing the input tracklets for the prediction module, some studies combine the tasks of tracking and prediction with joint optimization. PTP \cite{weng2021ptp} and PnPNet \cite{liang2020pnpnet} uses the shared feature representation to address both tasks. AffinPred \cite{weng2022whose}, TTFD \cite{zhang2023towards} use affinity matrices rather than tracklets as inputs of the prediction module to improve the forecasting performance, but they sacrifice the capability to provide tracking results explicitly. However, almost all of these methods are performed in a snap-shot form and neglect the misalignment issue between tracking and prediction. Compared to those approaches, our method uses a streaming framework and a unified spatio-temporal positional encoding method to address the above problems.


\section{APPROACH}

\subsection{Streaming Framework}
Simply, let $\mathcal{D} = \{d_1, \ldots, d_N\}$ represent the set of objects perceived in current frame from a 3D object detector, where $N$ denotes the number of objects. Concretely, each object at frame $t$ is represented as $d_i^t = [d_i^{\text{pos}, t}, d_i^{\text{size}, t}, d_i^{\text{head}, t}, d_i^{\text{class}, t}, d_i^{\text{score}, t}]$ where each element denotes the position, size, heading angle, class and confidence score from the module of detection, respectively. In this paper, the goal of joint 3D multi-object tracking and trajectory prediction includes two parts, to obtain the association of multiple obstacles in adjacent frames by assigning a unique track ID to each object, and meanwhile to predict the future trajectories $\mathcal{F} = \{f_1, \ldots, f_N\}$ for all agents in current frame, with each element of a trajectory specified by a two-dimensional coordinate $(x, y)$.


Based on the observation that the actual physical world is continuous and long-term history is essential for a safer autonomous driving system, we model the task of joint 3D multi-object tracking and trajectory prediction in a streaming manner (shown as Fig. \ref{fig:models}).  First of all, we extend the pattern of training from single-frame to multi-frame so as to narrow the gap between training and actual deployment. To be more specific, we introduce a Memory Bank for tracked objects to maintain long-term latent features for utilizing the long-term information more effectively, where the latent features are maintained through consecutive frames and could further benefit the performance of both tasks, including not only multi-object tracking but also trajectory prediction. 

Specifically, the memory bank consists of $ F \times N$ latent features where $F$ is the length of the memory bank and $N$ is the number of objects stored per frame. At each time, the latent feature of those tracked objects that have been associated with the new perceived objects in the current frame would be saved into the memory bank. These features are then utilized in subsequent frames to enhance features for tracked objects, detailed in Sec. \ref{sec:feature_interaction}. The entrance and exit of the memory bank follow the first-in, first-out rule.

\subsection{Spatio-Temporal Encoder}
\label{sec:feature_interaction}

\textbf{Feature Extraction.} To capture the semantic and motion information of the obstacles in the driving scenario efficiently and adequately, we conduct feature extraction for the tracked and new-come objects separately. For the perceived objects from adjacent frames, we use $d \in \mathbb{R} ^ {N_p \times C}$ and $\tau \in \mathbb{R} ^ {N_t \times C} $ to represent the semantic features, where $N_p$ and $N_t$ denote the number of objects at current frame $t$ (named as \textit{proposals}) and previous frame $t-1$ (named as \textit{tracklets}), respectively. At the same time, the historical trajectories of last $T_h$ frames for $N_t$ tracked objects are represented with $H \in \mathbb{R} ^ {N_t \times T_h \times C}$. Simply and effectively, we deploy the Multi-Layer Perceptron (MLP) to encode the semantic information into high-dimension features and fuse the historical data $H$ to \textit{trakclets} $\tau$ through a Multi-Head Cross Attention (MHCA) as:
\begin{equation}
    \label{eq:encoder}
    F_d = \operatorname{MLP}(d), \tilde{F_t} = \operatorname{MLP}(\tau) + \operatorname{MHCA}(\operatorname{MLP}(H))
\end{equation}
where $\tilde{F_t} \in \mathbb{R} ^ {N_t \times D}$, $F_p \in \mathbb{R} ^ {N_p \times D}$, and $C$, $D$ correspond to the dimension of the semantic and latent high-dimension features respectively.



Additionally, to equip our model with long-term temporal modeling capability, we exploit the latent features saved in the memory bank. Inspired by dynamic weight learning \cite{wang2023exploring} \cite{aydemir2023adapt}, an ego transformation is applied to ensure the temporal alignment and effective feature usage across frames:

\begin{equation}
    \label{eq:mln}
    \begin{aligned}
        \alpha, \beta  & = \operatorname{MLP} (E_{t} - E_{s}) \\
        M & =  \alpha \operatorname{LN} (\tilde{M}) + \beta  
    \end{aligned}
\end{equation}
where Eq.\ref{eq:mln} is an affine transformation and its parameters are derived from the ego difference between two frames. Then we apply temporal aggregation of long-term latent memory maintained in the memory bank for each tracked object with $\operatorname{MHCA}$ and then fuse the latent memory feature with the extracted feature of \textit{tracklets} as follows:
\begin{equation}
\label{eq: memory_encoder}
   F_t = \tilde {F_t} + \operatorname{MHCA}(M)
\end{equation}

\textbf{Spatio-Temporal Positional Encoding.} For the task of tracking, aligning all features within a unified coordinate system is essential for feature association. In contrast, for prediction tasks, previous research \cite{varadarajan2022multipath++}\cite{zhou2022hivt}  have demonstrated the advantages of agent-centric representations, which normalize various trajectories to local coordinate systems centered on the selected agent. To bridge the gap between coordinate representation between tracking and prediction, we propose a relative \textbf{S}patio-\textbf{T}emporal \textbf{P}ositional \textbf{E}ncoding (\textbf{STPE}) strategy. This approach differentiates between coordinate-independent and dependent features, using the former as query tokens for attention mechanism during feature interaction, while the latter is incorporated into attention through relative positional encoding.

To be specific, we encode the relative spatio-temporal position between object $i$ in previous frame $t$ (\textit{tracklet frame}) and object $j$ in current frame  $p$ (\textit{proposal frame}) as follows:  

\begin{equation}
    \label{eq:stpe_encode}
    \delta_{ij}^{tp} = \operatorname{MLP}([p_j^{p} - p_i^{t},  \theta_j^{p} - \theta_i^{t}] )
\end{equation}



\textbf{Attentional Spatio-Temporal Interaction.} Based on relative embedding from the spatio-temporal positional encoding strategy, we fuse the features of \textit{proposals} and \textit{tracklets} with cross-attention and self-attention iteratively. Take the proposal branch as an example, we use query-centric attention with a spatio-temporal positional encoding strategy, incorporating the relative positional embedding into key/value of the attention mechanism:

\begin{equation}
\label{eq:stpe_attn}
\begin{aligned}
F_i^{p\prime} &= \operatorname{MHCA}\left( 
\mathbf{Q} = F_i^{p}, \mathbf{K/V} = \{F_j^{t} + \delta_{ij}^{tp}\}_{j \in N_i}\right) \\
F_i^{p\prime} &= \operatorname{MHSA}\left( 
\mathbf{Q} = F_i^{p}, \mathbf{K/V} = \{F_j^{p} + \delta_{ij}^{p}\}_{j \in N_i}
\right) 
\end{aligned}
\end{equation}

As shown in Eq. \ref{eq:stpe_attn}, we first employ cross-attention to fuse the tracked objects' information from previous frames into new-come objects from current frames. Subsequently, self-attention is utilized within the current frame to foster awareness among detected objects in this frame. This process enables the features of newly perceived objects to incrementally assimilate comprehensive information, enriching their contextual awareness. We denote the result of this branch as \textit{proposals} context feature $F_i^{p'}$. Similarly, The tracklet branch undergoes the same propagation and gets \textit{tracklets} context feature $F_i^{t'}$ in parallel.

\subsection{MOT Head}

\textbf{Association with Optimal Transport.} The core purpose of the MOT head for StreamMOTP is to associate the M-tracked objects in the previous frame and the N-perceived objects in the current frame. To find the association relationship, we learn an affinity matrix $A^{\left ( \text{tp} \right )} \in \mathbb{R} ^ {N_t \times N_p}$ based on the \textit{tracklets} context feature and \textit{proposals} context feature after feature interaction. We use Dot Product to calculate the similarity pair, so each entry $A^{\left ( \text{tp} \right )}_{ij}$ represents the similarity score between the tracked object $i$ and the detected object $j$.

\begin{equation}
    A^{\left ( \text{tp} \right )}_{ij} = \frac{ \langle F^{t'}_i, F^{p'}_j \rangle }{\sqrt D}
, \forall (i, j) \in N_t \times N_p 
\end{equation}
where $D$ is the dimension of the context feature. 

Given the affinity matrix, we get the optimal affinity matrix $A^{(\text{opt})} \in \mathbb{R} ^ {(N_t +1) \times (N_p +1)}$ through \textbf{log sinkhorn algorithm} as SuperGlue \cite{sarlin2020superglue}, which performs differentiable optimal transport in log-space for stability. Under our streaming framework, the use of the log sinkhorn algorithm allows the model to modify the model parameters of previous frames while optimizing subsequent frames for continuous tracking and prediction. The last row and the last column of $A^{(\text{opt})}$ respectively represent newly appeared objects and tracklets without corresponding matched objects.

\textbf{Tracking Loss.} We supervise the output affinity matrix $A^{\text{(opt)}}$ with the ground truth (GT) relationship represented by the matrix $A^{\text{(g)}} \in \mathbb{R} ^ {(N_t +1) \times (N_p +1)}$. The accuracy of $A^{\text{(opt)}}$ is judged by how closely its high-value elements align with the ones in $A^{\text{(g)}}$. Therefore, we use the following loss:


\begin{equation}
    \mathcal{L}_{\text {tracking}} = -\frac{1}{N_m} \cdot (A^{\operatorname{(opt)}} e^{-U} + U) \cdot A^{(\text{g})}
\end{equation}
where the uncertainty matrix $U \in \mathbb{R} ^ {(N_t +1) \times (N_p +1)}$ is derived from \textit{tracklets} and \textit{proposals} feature to ensure the robustness of training, and $N_m$ is the number of matching pairs in $A^{\text(g)}$. Finally, we get association relationship $A$ from $A^{\text{(opt)}}$.

\subsection{Dual-Stream Predictor}
\label{approach: prediction}

\begin{figure}[t]
\centering
\includegraphics[scale=0.55]{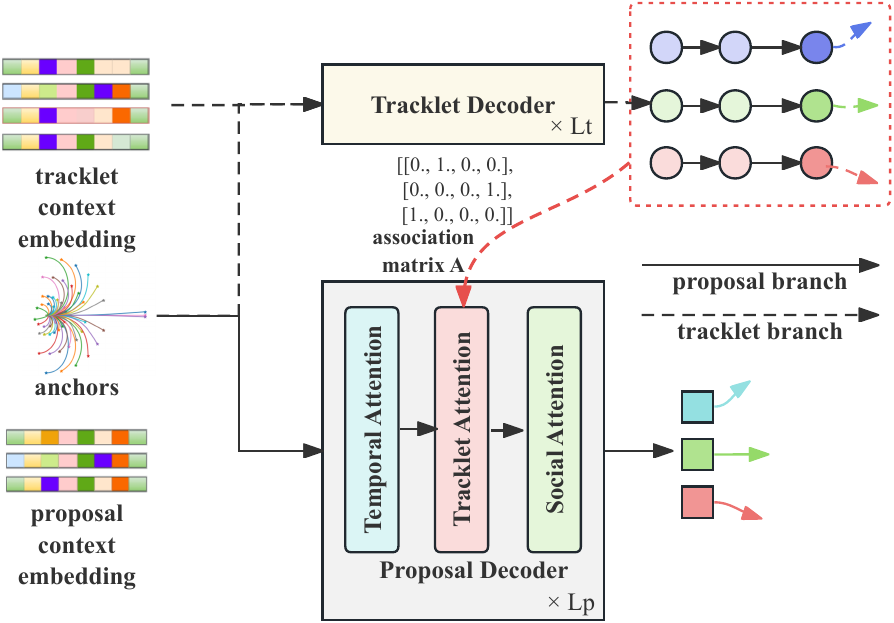}
\caption{Overview of dual-stream predictor. Two branches predict the previous frame trajectories and the current frame detections simultaneously, The streaming connection between consecutive frames smooth the predicted trajectories.}
\vspace{-15pt}
\label{fig:prediction}
\end{figure}

The predictor predicts all agents' multi-modal future trajectories. The detail of the predictor is shown in Fig. \ref{fig:prediction}.

\textbf{Single Frame Prediction}. To jointly predict all future trajectories for perceived objects in the current frame, we utilize a transformer-based decoder that incorporates the previous encoded context feature by learnable intention queries. To combine the advantages of the prior acceleration of convergence provided by the anchor-based model\cite{shi2022motion} and the high flexibility of the anchor-free model\cite{varadarajan2022multipath++}, we combine learnable tokens and anchors to form the query:

\begin{equation}
    Q_p^{l} = I + \phi(A_{T}) + \phi(\hat{x}_T^{l-1})
\end{equation}
where $Q_p^{l} \in \mathbb{R} ^ {N_p \times K \times D}$ is the query input at the current frame and layer $l$ decoder, which is combined from a learnable embedding $I$, the endpoints of the anchors $A_{T}$, and the predicted endpoints of previous layer $\hat{x}_T^{l-1}$, which are fused through $\phi$ (a sinusoidal position encoding followed by an $\operatorname{MLP}$). Next, to aggregate features from context embedding, we perform attention mechanism on the temporal and social dimensions to get multi-modal prediction output. 

\begin{figure}[t]
\centering
\includegraphics[scale=0.5]{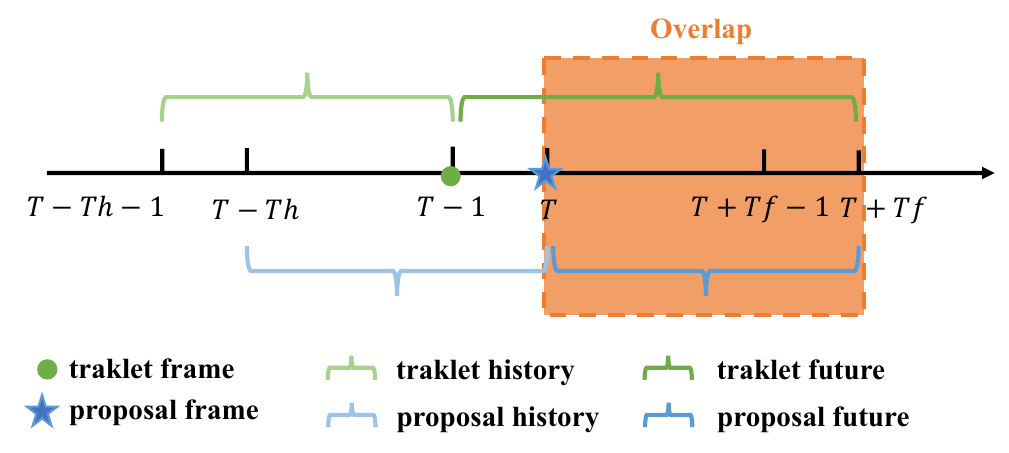}
\caption{The idea of temporal consistency between consecutive frames, where the consistency of the overlap is beneficial for aligning trajectories for continuity and stability.}
\vspace{-20pt}
\label{fig:tc}
\end{figure}

\textbf{Dual-Stream Predictor.} It is obvious that the predictions for previously tracked objects and currently perceived objects share a large overlap on those matched objects. As shown in Fig.\ref{fig:tc}, the $T_f$+1 predictions from frame $t$-1 should be consistent with the $T_f$ predictions from current frame $t$ in the last $T_f$ frames. And it's much more feasible to generate consecutive output trajectories with the streaming nature of the proposed framework of StreamMOTP.

Based on the observations, we propose a dual-stream predictor to improve the quality and temporal consistency of the predicted trajectories. The predictor comprises two branches: a primary branch focuses on making predictions for the detected objects in current frame and a supportive auxiliary branch for the previous tracked objects. The primary branch follows \textit{Single Frame Prediction} to predict from the context features of \textit{proposals}, while the auxiliary branch leverages the context features of \textit{tracklets} to generate $K$ adaptive predictions $\hat{Y_t} \in \mathbb{R} ^ {N_t \times K \times (T_f+1) \times 2}$ specific to the tracked objects. Since the prediction result $\hat{Y_t}$ from the \textit{tracklet frame} and $\hat{Y}_p$ from the \textit{proposal frame} have $T_f$ overlapping, using $\hat{Y_t}$ to guide the prediction of $\hat{Y_p}$ enhances both accuracy and temporal coherence of the predicted trajectory. Specifically, we encode and map the overlapping $T_f$ frame of $\hat{Y_t}$ to yield auxiliary features $F_{Y_t} \in \mathbb{R} ^ {N_p \times K \times T_f \times D}$:

\begin{equation}
    F_{Y_t} = \operatorname{MLP} (\operatorname{PE} (A^T \hat{Y_t})) 
\end{equation}
where $\operatorname{PE}(\cdot)$ denotes sinusoidal position encoding, $A \in \mathbb{R} ^ {N_t \times N_p}$ denotes the association matrix given by MOT head. 

In addition to \textit{Single Frame Prediction}, auxiliary features and anchor queries from the current frame are aggregated together in our dual-stream predictor. We adopt multi-head cross attention, taking the anchor embedding from the current frame as query, and the prediction features from the auxiliary tracklet branch as key and value:

\begin{equation}
    \label{eq:aux_attn}
    Q =  \operatorname{MHCA} (\mathbf{Q} = Q, \mathbf{K/V}=F_{Y_t})
\end{equation}

We place Eq. \ref{eq:aux_attn} after the interaction between queries and \textit{proposals} context features, while before the self-attention of the queries, making the queries interact sequentially with historical features, future features, and the social context.

\textbf{Multi-modal Prediction with Gaussian Mixture Model.} As the future behaviors of the agents are highly multi-modal, we follow \cite{varadarajan2022multipath++} to represent the distribution of predicted trajectories with Gaussian Mixture Model (GMM):

\begin{equation}
    \label{eq:gmm}
   f\left(\left\{\mathbf{Y}_{i}^{t}\right\}_{t=1}^{T_f}\right)=\sum_{h=1}^{K} p_{i, k} \prod_{t=1}^{T_f} \text { GMM }\left(\mathbf{Y}_{i}^{t} \mid \boldsymbol{\mu}_{i, k}^{t}, \mathbf{\sigma }_{i, k}^{t}\right) 
\end{equation}
where $\left\{p_{i, k}\right\}_{k=1}^{K}$ is the probability distribution between $K$ modes, and the $k-$th mixture component’s Gaussian density for agent $i$ at time step $t$ is parameterized by $\mu_{i,k}^{t}$ and $\sigma_{i,k}^{t}$. Given Eq. \ref{eq:gmm} for all predicted steps, we adopt negative log-likelihood loss and supervised predictions for new-come objects in the current frame and predictions for the tracked objects simultaneously. Loss can be formulated as:

\begin{equation}
    \label{eq:nll}
    \mathcal{L}_{\text {prediction}} = - \log f(\hat{Y_p}) - \log f(\hat{Y_t})
\end{equation}

Then, the final loss of our model is denoted as:

\begin{equation}
\label{eq:total_loss}
\mathcal{L} = \lambda \mathcal{L}_{\text {tracking}} + \mathcal{L}_{\text {prediction}}
\end{equation}
where $\lambda \in \mathbb{R}_{>0}$ is the weight for tracking loss to balance the the joint optimization of the two tasks.


\section{EXPERIMENTS}

\begin{table*}[!ht]
    \centering
    \caption{Comparison with existing approaches (on nuScenes). All results is based on detections from Megvii.}
    \begin{subtable}[t]{0.31\linewidth}
    \centering                
    \caption{3D MOT Performance}
     \scalebox{1}{
    \begin{tabular}[t]{@{}ccc@{}}
    \toprule
    Methods & AMOTA $\uparrow$ & MOTA $\uparrow$ \\
    \midrule
    mmMOT \cite{zhang2019robust} & 23.93 & 19.82 \\
    GNN3DMOT \cite{weng2020gnn3dmot} & 29.84 & 23.53 \\
    AB3DMOT \cite{weng2020ab3dmot} & 39.90 & 31.40 \\
    PTP \cite{weng2021ptp} & 42.36 & 32.06 \\ 
    \midrule
    StreamMOTP & \textbf{46.30} & \textbf{40.50} \\
    \bottomrule
    \end{tabular}}
      
      \label{subtab:tracking} 
    \end{subtable}
\hfill             
    \begin{subtable}[t]{0.31\linewidth}
 \centering
 \caption{One Step MOTP Performance}
 \scalebox{1}{
    \begin{tabular}[t]{@{}ccc@{}}
        \toprule
        Methods & minADE $\downarrow$ & minFDE $\downarrow$ \\
        \midrule
        Social-GAN \cite{gupta2018social} & 1.794 & 2.850 \\ 
        TraPHic \cite{chandra2019traphic}& 1.827 & 2.760 \\
        Graph-LSTM \cite{chandra2020forecasting} & 1.646 & 2.445 \\
        PTP \cite{weng2021ptp} & 1.017 & 1.527 \\
        \midrule
        StreamMOTP & \textbf{0.810} & \textbf{1.481} \\
        \bottomrule
    \end{tabular}}
      
      \label{subtab:ptp}
    \end{subtable}%
\hfill
    \begin{subtable}[t]{0.31\linewidth}
     \centering
     \caption{Multi Step MOTP Performance}
     \scalebox{1}{
        \begin{tabular}[t]{@{}ccc@{}}
        \toprule
        Methods & minADE $\downarrow$ & minFDE $\downarrow$ \\
        \midrule
        PTP \cite{weng2021ptp} & 2.320 & 3.819 \\ 
        MTP(S=10) \cite{weng2021ptp} & 1.585 & 2.512 \\
        MTP(S=200) & 1.325 & 1.979 \\
        AffinPred \cite{weng2022whose} & 0.977 & 1.628\\
        \midrule
        StreamMOTP & \textbf{0.757} & \textbf{1.487} \\
        \bottomrule
        \end{tabular}}
    \label{subtab:affinpred}
    \end{subtable}%

    \label{tb:overall}
    \end{table*}


\begin{table*}
 \centering
 \caption{Ablation study on the components of StreamMOTP.}
 \setlength{\tabcolsep}{8pt} 
 \scalebox{1.1}{
    \begin{tabular}[t]{@{}ccc|ccc|cccc@{}}
    \toprule
    Memory Bank & STPE  & Stream Predictor & AMOTA & AMOTP & MOTA & minADE & minFDE & MR & tc\\
    \midrule
      &      &      &  0.523     &   0.781    &  0.426    &   0.572     & 0.942  &  0.113  &  - \\	    
     &   \checkmark   &    \checkmark  & 0.556     &   0.770    &  0.466    &   0.384     & 0.594  & 0.075  &  - \\	
     \checkmark  &    &    \checkmark  &  0.528     &   0.782    &  0.431    &   0.524     & 0.838  &  0.103  &  - \\	
    \checkmark   &   \checkmark   &    &  0.544    &   \textbf{0.768}    &  0.456    &   0.488     & 0.776  &  0.098  &  2.081 \\	
    \checkmark   &   \checkmark   &    \checkmark  &  \textbf{0.556}     &   0.779    &  \textbf{0.472}    &   \textbf{0.377}     & \textbf{0.586}  &  \textbf{0.072}  &  \textbf{1.942} \\									
    \bottomrule
    \end{tabular}}
\label{tb:ablation}
\vspace{-10pt}
\end{table*}%

\subsection{Experimental Setup and Implementation Details}

\textbf{Dataset and Metrics.} The proposed method is evaluated on the popular nuScenes dataset. Following the standard practices \cite{caesar2020nuscenes} of nuSences dataset, we predict trajectories for objects perceived in the current frame and use the distance threshold of 2m to match them with GT future trajectories. In the task of trajectory prediction, the models predict future trajectories for 3s and 6s to align with other works, with a time interval of 0.5s, based on 2s historical data. As for the task of MOT, We employ the commonly-used AMOTA, MOTA, and AMOTP for evaluation. And standard minADE and minFDE metrics are used to evaluate the prediction performance. Moreover, we design the metric of ‘\textbf{tc}’ to evaluate the temporal consistency, which is calculated as the ADE in $T_f-1$ overlapping frames between predictions from $T$ to $T+T_f$ and predictions from $T-1$ to $T+ T_f -1$.



\textbf{Inputs.} In StreamMOTP, input data is formatted in a sequential format. During training, we split the streaming video into training slices and use a sliding window to sequentially get the inputs at each timestamp. To address detector noise, we incorporate the detected results and employ the ground truth (GT) matching relationships up to the $(t-1)$-th frame to create history tracks. Newly perceived objects without association in the current frame serve as \textit{proposals}. In online inference, the model takes raw detections as input to perform tracking and prediction jointly.

\textbf{Training.} To avoid poor latent memories which may impede the training procedure in early stages, \textbf{scheduled sampling} \cite{bengio2015scheduled} is applied to the memory bank. We train our model for 180 epochs. Specifically, features in the memory bank are selected through sampling, and the sampling rate starts to increase at epoch 30, following a sigmoid curve. 


\subsection{Comparison with Related Work}

Table \ref{tb:overall} compares StreamMOTP with other methods in tracking and prediction, using the same Megvii\cite{zhu2019class} detector for fairness. For MOT, we evaluate all categories, while for trajectory prediction, we adopt two settings from prior studies:  (1) \textbf{Setting1: One Step MOTP}. In Setting1, we follow a single-step tracking and 3s prediction, similar to PTP \cite{weng2021ptp}. The model uses GT past trajectories $ t \in \{T_c-T_h, \cdots,  T_c-1\}$ and GT detections in the current frame $T_c$, conducts MOT at the current frame, and forecasts future trajectories in frames $ t \in \{T_c + 1, \cdots T_c + T_f\}$. Results for all classes from the nuScenes Prediction Challenge are reported. This setting is more suitable for Vehicle-to-Vehicle (V2V) scenario.  (2) \textbf{Setting2: Multi Step MOTP}. In setting2, we perform standard tracking and 6s prediction for detected objects in $T_c$, based on their tracked histories, and evaluate prediction results on all vehicle classes. This setting aligns more closely with the current stage of autonomous driving and is more widely adopted in industry deployments.

Our model surpasses previous related work in both tasks of multi-object tracking and trajectory prediction. In MOT performance, shown in Table \ref{subtab:tracking}, our model not only achieves gains over PTP baseline \cite{weng2021ptp} with improvements of 3.94\% in AMOTA and 8.44\% in MOTA,  but also surpasses several competing trackers. Table \ref{subtab:ptp} shows the prediction comparison for one-step MOTP. Our model reaches the lowest minADE of 0.810 and minFDE of 1.481, which outperforms PTP \cite{weng2021ptp} by 0.207 on minADE and 0.046. Moreover, Table \ref{subtab:affinpred} offers a comparison of multi-step MOTP's predictions, where our model attains state-of-the-art performance with a minADE of 0.757 and a minFDE of 1.487, outperforming AffinPred \cite{weng2022whose} by 0.220 and 0.141, respectively. The improvements in Table \ref{subtab:affinpred} are more obvious than in Table \ref{subtab:ptp} for the reason that trajectory prediction in setting1 is more saturated than in setting2, indicating the larger growth potential for prediction based on tracked trajectory.


\begin{figure*}[t]
\centering
\includegraphics[scale=0.42]{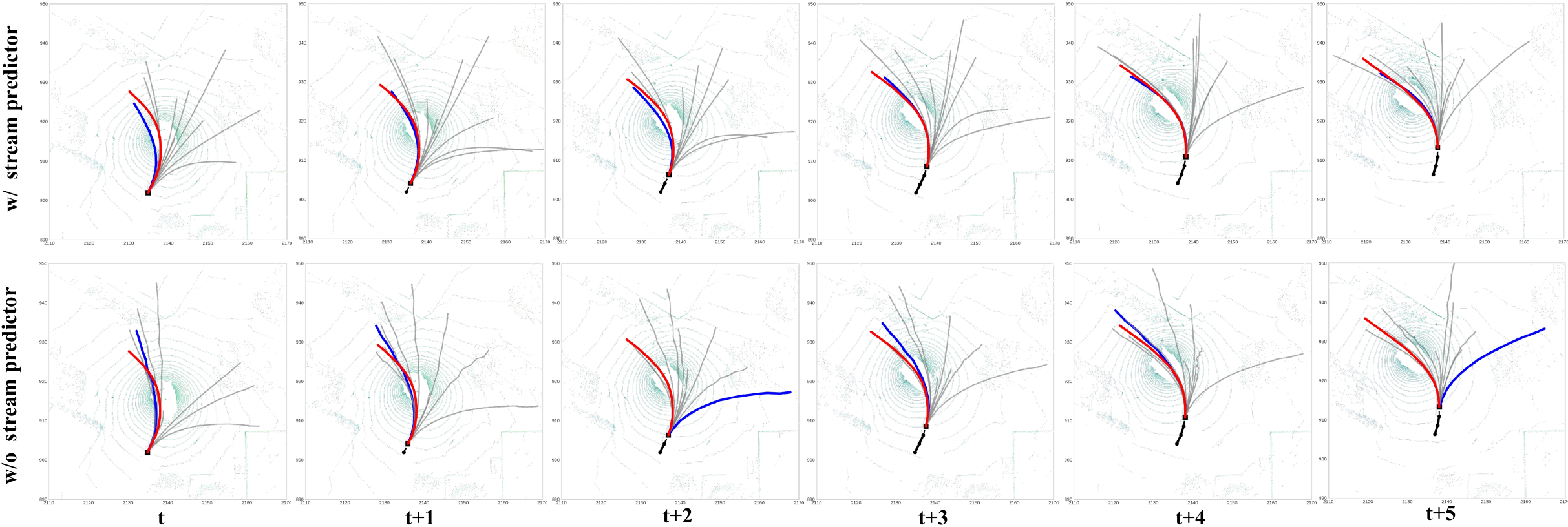}
\caption{Qualitative results of StreamMOTP on the nuScenes validation set during consecutive frames. The tracked history and detection are shown in black, models' best score prediction and ground-truth trajectories are drawn in blue and red respectively. The predictions of other modes are drawn in gray. The top row shows the results given by the dual-stream predictor while the bottom row shows the results with a base predictor.}
\vspace{-15pt}
\label{fig:res1}
\end{figure*}

\begin{table}[t]
 \centering
 \caption{The effect of training slice length (Abbreviated as "Slice") and memory bank (Abbreviated as "Mem").}
 \vspace{-5pt}
 \scalebox{0.95}{
    \begin{tabular}{c|c|cc|ccc}
    \toprule
    Slice & Mem & AMOTA & MOTA  & minADE & minFDE & MR     \\
    \midrule
    3            &        & 0.570 & 0.490 & 0.633 & 0.953 & 0.137  \\
    5            &        & 0.560 & 0.478 & 0.402 & 0.621 & 0.075 \\
    10           &        & 0.557 & 0.466 & 0.384 & 0.594 & 0.075 \\
    \midrule
    3            &\checkmark  & 0.570  & 0.486 & 0.537 & 0.813 & 0.119 \\
    5            & \checkmark & 0.564 & 0.478 & 0.392 & 0.602 & 0.072 \\
    10           & \checkmark & 0.556 & 0.472 & \textbf{0.377} & \textbf{0.586} & \textbf{0.072} \\
    \bottomrule
    \end{tabular}}
\label{tb:memory}
\vspace{-8pt}
\end{table}%

\begin{table}[t]
 \centering
 \caption{Ablation study of Memory Bank in Slice=3.}
 \vspace{-5pt}
 \scalebox{0.95}{
    \begin{tabular}[t]{@{}c|cc|ccc@{}}
    \toprule
     Memory Length & AMOTA & MOTA & minADE & minFDE & MR \\
    \midrule
     0 & 0.570 &   0.490   &    0.633  &  0.953    &  0.137   \\
     1 & 0.569 &  0.487    &    0.603  &  0.921    &  0.135   \\
     2 & 0.570 &   0.486   &    \textbf{0.537}  &  \textbf{0.813}    &  \textbf{0.120}   \\
    \bottomrule
    \end{tabular}}
\label{tb:bnk}
\vspace{-20pt}
\end{table}

\subsection{Ablation Studies}

We evaluated the impact of each module within our StreamMOTP framework, as summarized in Tabel \ref{tb:ablation}, where the bottom row represents the full implementation of our method. All models are experimented on Setting2, except that the detector is switched to CenterPoint\cite{yin2021center} and 3s prediction metrics are computed on True Positive detections at a recall rate of 0.6. The Megvii detector, being an older model, exhibits subpar detection capabilities. Therefore, we switch to a detector with relatively moderate performance to better measure each module's efficacy.

\textbf{Effects of each module.} Firstly, upon removing the memory bank, we observed a slight decline in performance for both tracking and prediction tasks. We will further explore its impact later. Secondly, we remove the spatio-temporal positional encoding in the spatio-temporal interaction module and encode the absolute coordinate feature in the same way as the attribute feature. There is a significant drop in performance for both tasks of tracking and prediction, which shows that spatio-temporal positional encoding maintains the pose-invariance for trajectory predictions and effectively addresses the issue of inconsistent coordinate representations. Thirdly, we replace the dual streaming predictor with a single frame predictor performed only on the current frame. The second-last row shows that the dual-stream predictor plays a vital role in advancing prediction performance. The modest decrease in tracking further corroborates that augmenting prediction capabilities also benefits tracking results. Notably, the tc metric also drops when the dual-stream predictor is eliminated, which indicates that the dual-stream predictor enhances the trajectory predictions' quality and consistency. The reason is that in two consecutive frames, predictions from previous frames serve as a valuable prior reference for predicting current perceived objects' trajectories, which helps to yield more viable and steady outcomes.


\textbf{Effects of streaming framework.} The effectiveness of the streaming framework and the memory bank is explored by adjusting the lengths of training segments. In Table \ref{tb:memory}, tracking performance stays consistent, whereas prediction accuracy significantly benefits from longer training slices due to its dependence on sequential and extensive sequential information. This finding stems from the gap that our models are trained in split slices (multi-frame sequences of length $k$) but evaluated in streaming video (the average length is 40 in nuScenes, $k \ll 40$). This gap constrains the effectiveness of approaches, especially for previous snap-shot methods. Our streaming framework narrows this gap between the segmented training approach and continuous video inference by utilizing temporal information over successive frames, thus enhancing prediction performance. Moreover, the integration of the memory bank, particularly with shorter slices, markedly boosts prediction accuracy by the retention and utilization of long-term latent features in the memory bank, therefore improving the model's capability for long-term sequence modeling. This is crucial under resource constraints that limit slice length and temporal receptive field. Furthermore, Table \ref{tb:bnk} shows that as the length of the memory bank expands, the model's performance grows, which further demonstrates the impact of the memory bank.


\begin{table}[t]
 \centering
 \caption{Model performance on varying detectors.}
 \scalebox{0.9}{
\begin{tabular}{@{}c|ccc|ccc@{}}
\toprule
Detectors   & AMOTA & AMOTP & MOTA  & minADE & minFDE & MR    \\
\midrule
Megvii      & 0.463 & 0.997 & 0.405 & 0.470  & 0.751  & 0.096 \\
CenterPoint & \textbf{0.556} & \textbf{ 0.779} & \textbf{0.472} & \textbf{0.377}  & \textbf{0.586}  &\textbf{ 0.072}
\\
\bottomrule
\end{tabular}
}
\label{tb:detector}
\vspace{-15pt}
\end{table}

\textbf{Generalization performance on different detectors.} We applied our model with different detectors and summarized the result in Table\ref{tb:detector}. The significant growth of CenterPoint compared to Megvii in tracking and 3s prediction underscores our model's strong generalization ability, independent of specific detectors. It is anticipated that the model will achieve superior performance with advanced detectors.

\subsection{Qualitative Results}

We provide some qualitative results in Fig. \ref{fig:res1} to show our predictions. There is a brand new object without historical trajectory perceived at frame $t$. StreamMOTP successfully predicts its future trajectory with social interactions. Moreover, by comparing the two rows, we can see that all mode predictions in the top row are smoother and more precise, and the highest score of the predictions fluctuates less.

\section{CONCLUSIONS}



In this paper, we introduce StreamMOTP, a streaming and unified framework for joint multi-object tracking and trajectory prediction. With the design of the memory bank, spatio-temporal positional encoding strategy, and dual-stream predictor, streamMOTP bridges the gap between training and actual deployment, as well as maintains better capability and great potential for both tasks of multi-object tracking and trajectory prediction. The experiments on nuSences demonstrate the effectiveness and superiority of the proposed framework. We hope this work could further offer insights into the multi-task end-to-end autonomous driving systems.

\bibliographystyle{IEEEtran}
\bibliography{zjh_references}

\end{document}